\documentclass[letterpaper]{article} 
\usepackage{aaai25}  
\usepackage{times}  
\usepackage{helvet}  
\usepackage{courier}  
\usepackage[hyphens]{url}  
\usepackage{amsthm,amsmath,amssymb}
\usepackage{graphicx} 
\usepackage{natbib}  
\usepackage{caption} 
\usepackage{multirow}
\usepackage{makecell}
\usepackage{listings}
\usepackage{booktabs} 
\usepackage{diagbox}
\usepackage{subfigure}
\usepackage{colortbl}
\usepackage[table]{xcolor}
\usepackage{algorithm}
\usepackage{algorithmic}

\usepackage{newfloat}
\usepackage{listings}
\DeclareCaptionStyle{ruled}{labelfont=normalfont,labelsep=colon,strut=off} 
\floatstyle{ruled}
\newfloat{listing}{tb}{lst}{}
\floatname{listing}{Listing}
\title{StyleMark: A Robust Watermarking Method for Art Style Images Against Black-Box Arbitrary Style Transfer}
\author{
    Yunming Zhang\textsuperscript{\rm 1},
    Dengpan Ye\textsuperscript{\rm 1},
    Sipeng Shen\textsuperscript{\rm 1},
    Jun Wang\textsuperscript{\rm 2}
}
\affiliations{
    \textsuperscript{\rm 1}Key Laboratory of Aerospace Information Security and Trusted Computing, Ministry of Education. School of Cyber Science and Engineering, Wuhan University\\
    \textsuperscript{\rm 2}School of Computing and Information Technology, Great Bay University\\

}

\usepackage{bibentry}

\begin{document}

\maketitle

\begin{abstract}



Arbitrary Style Transfer (AST) achieves the rendering of real natural images into the painting styles of arbitrary art style images, promoting art communication. However, misuse of unauthorized art style images for AST may infringe on artists' copyrights. One countermeasure is robust watermarking, which tracks image propagation by embedding copyright watermarks into carriers. Unfortunately, AST-generated images lose the structural and semantic information of the original style image, hindering end-to-end robust tracking by watermarks. To fill this gap, we propose StyleMark, the first robust watermarking method for black-box AST, which can be seamlessly applied to art style images achieving precise attribution of artistic styles after AST. Specifically, we propose a new style watermark network that adjusts the mean activations of style features through multi-scale watermark embedding, thereby planting watermark traces into the shared style feature space of style images. Furthermore, we design a distribution squeeze loss, which constrain content statistical feature distortion, forcing the reconstruction network to focus on integrating style features with watermarks, thus optimizing the intrinsic watermark distribution. Finally, based on solid end-to-end training, StyleMark mitigates the optimization conflict between robustness and watermark invisibility through decoder fine-tuning under random noise. Experimental results demonstrate that StyleMark exhibits significant robustness against black-box AST and common pixel-level distortions, while also securely defending against malicious adaptive attacks. 

\end{abstract}

%

\section{Introduction}

\begin{figure}[!t]
\centering
\includegraphics[width=82mm]{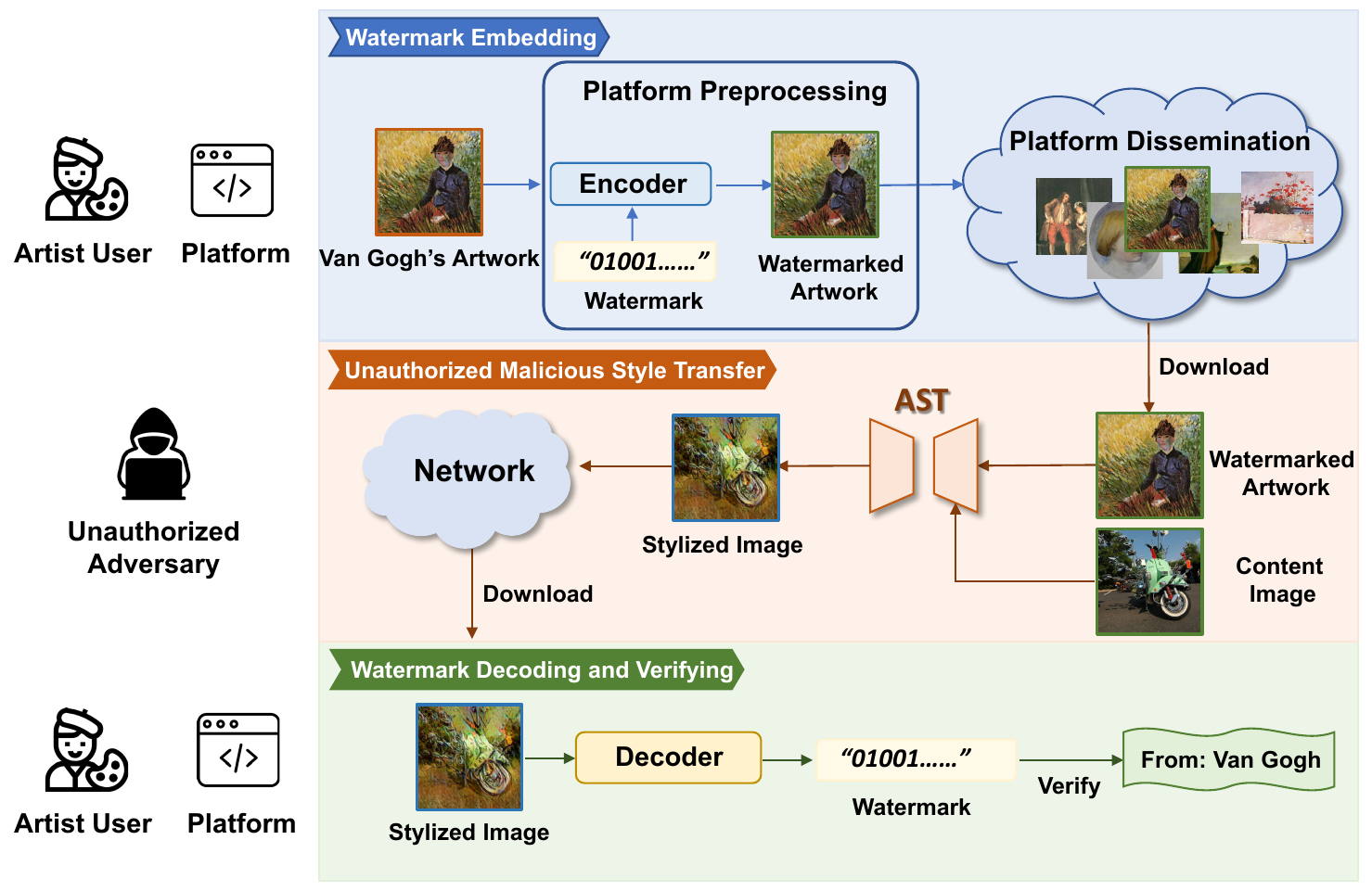}
\caption{Protection scenarios for StyleMark. Artists register on the art-sharing platform and upload their art style images. The platform embeds copyright watermarks before public release. Unauthorized adversaries may download the art style image and use AST for style transfer. When misuse is detected, the platform can extract the watermark from stylized images to verify the source of the art style.}
\label{fig_1}
\end{figure}

Artists express creative ideas and spiritual world through their paintings. Many artists share their artworks (such as hand drawings, illustrations, and original paintings) on public websites like ArtStation, Theispot, and PIXIV. Recently, Arbitrary Style Transfer (AST) algorithms have made significant advancements, enabling rapid style feature transfer from any art style to natural content images using flexible feedforward networks~\cite{zhang2022domain,hong2023aespa}. However, the rapid development of AST algorithms has introduced new challenges to the art-sharing community~\cite{guo2024artwork}. As illustrated in Figure~\ref{fig_1}, an unauthorized adversary can easily use AST to transform their natural content images into the style of an artist's style image, falsely claiming it as their own artistic style, thus leading to copyright disputes. In 2023, the social platform Little Red Book was sued by four artists for unauthorized commercial use of their paintings, generating stylized images using AST algorithms for profit~\cite{ThePaper}. One plaintiff stated that when copying a style becomes easy, no one will be motivated  to create new styles. Therefore, it is crucial to establish safeguards against the misuse of AST technology.

\begin{figure*}
	\centering
	\includegraphics[width=165mm]{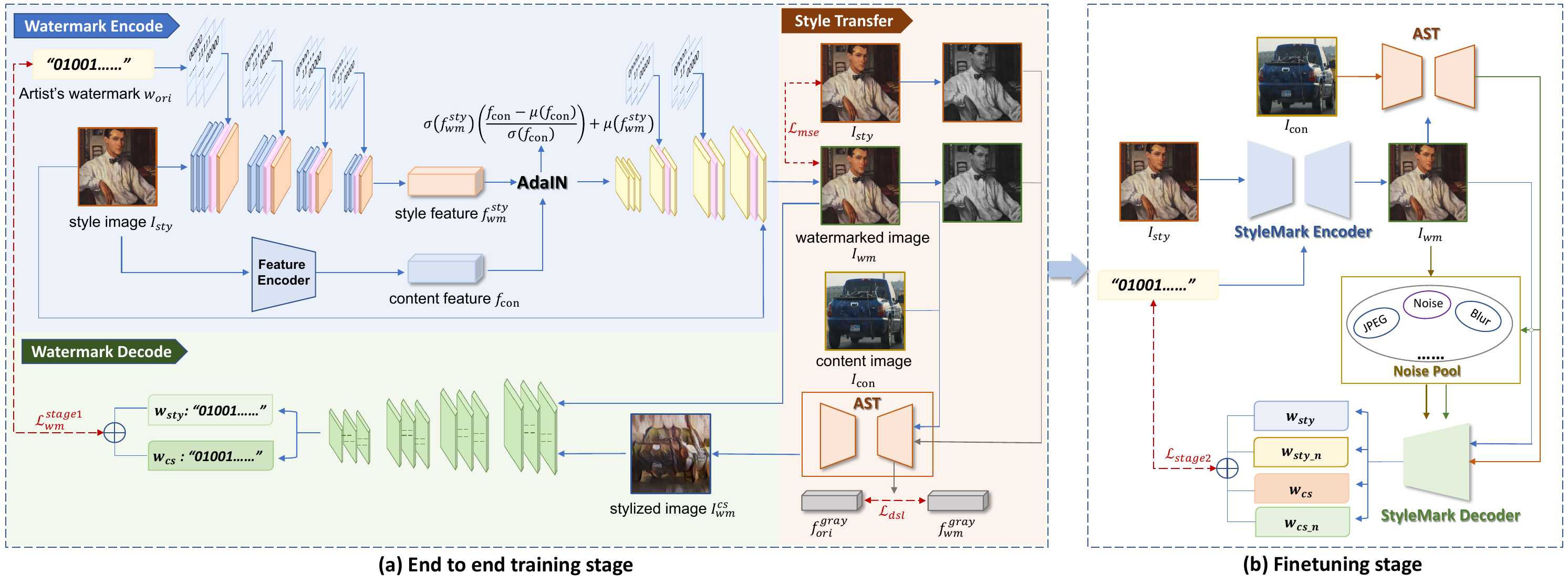}
        \caption{The whole pipeline of our StyleMark.}	
 \label{outline}
\end{figure*}

It has been investigated to counteract AST through adversarial attacks~\cite{li2024neural}, which disrupts AST by introducing adversarial perturbations to art style images. However, it introduces more obvious artifacts to artworks and only reduces the quality of AST-generated images, which still exhibit target style features to some extent. Especially for black-box models, the adversarial effect is significantly diminished. Most importantly, adversarial attack-based defense methods completely block the application of AST for artworks, including legitimate usage. This hinders artistic exchange and the commercial development of artworks.

To address the limitations of current defense methods, we propose a new concept inspired by deep watermarking technology, called ``StyleMark". This is the first watermark-based method for protecting the copyright of art style images against black-box AST. As shown in Figure~\ref{fig_1}, the art-sharing platform can use the StyleMark encoder to embed copyright watermarks into artworks, tracking their dissemination. After malicious AST, the StyleMark decoder can extract the watermark from generated stylized images to verify the original creators of the art style. 

Several deep robust watermarking algorithms have been proposed to enhance robustness against common pixel-level distortions and achieve precise image-level attribution~\cite{fang2023flow}. However, they are ineffective for style tracing in art images after AST. This is because the AST process differs from common image processing or local editing operations, as it generates stylized images retaining only high-level brushstroke features of art style images. The structural and semantic information is derived entirely from additional natural content images, severely disrupting the watermark distribution and interrupting end-to-end watermark tracking. In this paper, we propose a mild hypothesis and design StyleMark accordingly: \textbf{\textit{Watermark information can accompany the shared style features of art style images and be co-rendered into the generated images across different AST.}} To achieve this, we propose a novel style watermark network. By learning the feature encoding and reconstruction process of AST, the StyleMark encoder embeds watermarks into the multi-scale shared style feature space. It adjusts the neural activation distribution of style features through watermarks, enabling style feature shifts accompanied by the watermark traces. Additionally, we specifically design the distribution squeeze loss to regularize global image quality loss. It forces the model to learn to retain watermark information in redundant style features while mitigating the loss of structural details due to statistical feature replacement. Finally, based on end-to-end training, we fine-tuned the decoder with a random noise pool, ensuring the security of StyleMark against multiple post-processing attacks.

Our contribution can be summarized as follows:

\begin{itemize}

\item We innovatively propose StyleMark, the first plug-and-play robust watermarking algorithm designed for artwork copyright protection against black-box AST.



\item We present a new style watermark network. By embedding the watermark in the multi-scale shared feature space, we achieve watermark-accompanied style feature rendering across various AST processes.

\item We specifically design a distribution squeeze loss to optimize intrinsic watermark distribution and promote the recovery of structural details in watermarked images.


\item  We conduct extensive experiments on seven AST models and various pixel-level distortions, demonstrating StyleMark's remarkable black-box robustness and security against adaptive attacks. It surpasses other comparative methods by up to $43\%$ in watermark recovery accuracy.


\end{itemize}

\section{Related Works}

 
\subsection{Arbitrary Style Transfer} Style transfer is a non-photorealistic rendering technique closely related to texture synthesis.~\cite{huang2017arbitrary} introduce adaptive instance normalization to adjust the channel statistics of content features, enabling the first arbitrary style image rendering with a unified model~\cite{deng2020arbitrary,wu2022ccpl,zhang2022exact,mccstn}. Some studies have further developed this encoder-decoder framework by incorporating unique feature fusion or attention networks to achieve more natural content image rendering~\cite{park2019arbitrary}. To address the issue of fine-grained style feature loss in attention-based methods, ~\cite{mccstn} introduce a new style representation and transfer framework. This framework employs a feature fusion module to integrate the aesthetic features of style images with the fine-grained details of content images. ~\cite{cap} construct a novel AST model comprising reversible residual networks and unbiased linear transformation modules, preventing the loss of content affinity information during forward and backward inference processes. The rapid advancement of AST aids in expanding the influence and commercial value of arworks. However, its simplicity and efficiency raise concerns about the illicit misuse of artworks.

\subsection{Robust Watermarking} 

Deep learning has inspired the use of neural networks in robust watermarking algorithms, replacing handcrafted low-level features for greater resilience against advanced digital distortions~\cite{lan2023robust,liu2023detecting}. ~\cite{zhu2018hidden} propose the first robust watermarking framework based on DNNs, consisting of an encoder-decoder, discriminator, and noise layer for robust training. Subsequent research built upon this framework by enhancing algorithm robustness through the design of various simulated noise layers~\cite{jia2021mbrs}. Advancements in deep image editing models necessitate robust watermarking algorithms to resist tampering by local editing algorithms. Some methods enhance robustness against deep tampering models by integrating target editing models into simulated noise pools, but only primarily addressing localized modifications in the carrier~\cite{wang2021faketagger,hu2024robust}. Recent methods shift focus from image-level attribution to employing binary classifiers to detect whether the image contains a specific user watermark~\cite{pan2023anchmark,ma2024pigw}. These methods reduce robustness requirements but limit wider applications. The uniqueness of artworks necessitates watermark methods for precise image-level style attribution. However, AST disrupts the structural and semantic information of the carrier style image, making current watermarking methods ineffective for style attribution after AST.



\section{Threat Model}

We consider a tripartite threat model based on the characteristics of artworks dissemination on the internet.

\subsection{Artist User}

Artists register their identities on the art-sharing platform and upload their artworks. They have the right to request copyright protection and verify art style attribution in case of misuse of artworks. Artists may also authorize free AST creations by commercial or online users to develop the commercial value of their artworks. Therefore, artists may prefer copyright protection that does not alter the appearance of their images or interfere with authorized AST applications.


\subsection{Art-Sharing Platform}

The platform backend deploys the StyleMark model. When the artist uploads their artworks $I_{sty}$, The platform uses StyleMark encoder $\mathcal{E}(\cdot)$ embeds an invisible watermark ${{w_{ori}}\in [0,1]}^{1\times l}$ representing the artist’s unique copyright identifier to obtain the watermarked style image: $I_{wm}=\mathcal{E}(I_{sty},w_{ori})$, which is then publicly shared. The watermark embedding does not affect the appearance or compliant use of artworks. Once the platform detects the misuse of artworks to generate stylized images through AST, it can use StyleMark decoder $\mathcal{D}(\cdot)$ extracting the original artist watermark from stylized imaged to verify the style source. Notably, the platform lacks any knowledge of adversary's AST techniques, including the underlying algorithms and models.


\subsection{Unauthorized Adversary}

Unauthorized adversaries are not authorized by the artist to use the artwork for derivative creations. However, they possess open source pre-trained AST models. After downloading art images from art-sharing platforms, they can easily use the AST model to transfer the artistic style of the unauthorized artwork to their own natural content images without any model fine-tuning to get the forged stylized images: $I_{cs}=\mathcal{AST}(I_{sty},I_{con})$. This stylized image may be used for illicit commercial gain.



\section{Methodology}

We outline the proposed StyleMark in Figure~\ref{outline} and detail the algorithmic process in Alg.~\ref{alg}. Below we provide a comprehensive description of each component.

\begin{algorithm}[tb]
\caption{StyleMark Training Framework}
\label{alg}
\textbf{Input}: 
$I_{sty}$ (style image); $I_{con}$ (content image); $w_{ori}$ (watermark); $\mathcal{E}(\cdot)$ (StyleMark encoder); $\mathcal{D}( \cdot )$ (StyleMark decoder); $\mathcal{AST}(\cdot)$ (AST model); $\mathcal{F}_{con}(\cdot)$ (content feature encoder); $\mathcal{NP}(\cdot)$ (noise pool); $rand$ (random selection).\\
\textbf{Output}: best model parameters;
\begin{algorithmic}[1] 
\STATE Initialize $\mathcal{E}( \cdot )$ and $\mathcal{D}( \cdot )$;  
\FOR{$i\in  [ 0,Stage1\_iter  ]$ }
\STATE $I_{wm}  \Leftarrow \mathcal{E}(I_{sty}, w_{id}) $;
\STATE $f_{ori}^{gray} ,f_{wm}^{gray} \Leftarrow \mathcal{F}_{con}(gray(I_{wm}),gray(I_{sty}))$;
\STATE $I_{wm}^{cs}  \Leftarrow \mathcal{AST}(I_{wm}, I_{con}) $;
\STATE $w_{s}, w_{cs} \Leftarrow \mathcal{D}(I_{wm},I_{wm}^{cs}) $;
\STATE Compute the first-stage total loss with Eq.(\ref{13});
\STATE Update $\mathcal{E}( \cdot )$ and $\mathcal{D}( \cdot )$;
\ENDFOR
\FOR{$i\in  [ Stage1\_iter,Stage2\_iter  ]$ }
\STATE Loading encoder and decoder parameters;
\STATE $I_{wm}  \Leftarrow \mathcal{E}(I_{sty}, w_{id}) $;
\STATE$ w_{s}^{np} \Leftarrow \mathcal{D} (rand(\mathcal{NP}(I_{wm})))$;
\STATE $I_{wm}^{cs}\Leftarrow \mathcal{AST} (I_{wm},I_{con})$;
\STATE $w_{cs}^{np}\Leftarrow \mathcal{D} (rand(\mathcal{NP}(I_{wm}^{cs})))$;
\STATE Compute the second-stage loss using Eq.(\ref{16}).
\STATE Update $\mathcal{D}( \cdot )$;
\ENDFOR
\STATE \textbf{return} best model parameters.
\end{algorithmic}
\end{algorithm}

\subsection{StyleMark Encoder Optimization}

\subsubsection{Watermark Encoding.}To enable co-rendering of watermark traces with style features during the AST, we intuitively aim to embed watermarks into the general style features of the style image, ensuring compatibility with various black-box AST models. Most AST methods rely on similar front-end feature decoupling modules to extract high-level features from both style and content images, followed by different model structures for feature alignment and reconstruction. Thus, we innovatively incorporate a general AST framework into the watermark encoding feedforward network, proposing a novel style watermark encoder $\mathcal{E}(\cdot)$ that guides style feature shifts through watermark information.

The StyleMark encoder consists of a style feature encoder $\mathcal{F}_{sty}(\cdot)$, a content feature encoder$\mathcal{F}_{con}(\cdot)$, and a feature reconstructor $\mathcal{F}_{dec}(\cdot)$. The encoder receives the style image and its copyright watermark ${{w_{ori}}\in [0,1]}^{1\times l}$. $\mathcal{F}_{sty}(\cdot)$ extracts high-level stroke features of the style image, with multi-scale stroke features $f_{sty}$  from intermediate layers fused with redundant watermark information. Simultaneously, the style image $I_{sty}$ is used as input for the content feature encoder $\mathcal{F}_{con}(\cdot)$ to extract content features $f_{con}$ for subsequent image reconstruction: $f_{con}= \mathcal{F}_{con}(I_{sty}).$



The binary watermark sequence is initially duplicated in the spatial dimension and then expanded in the channel dimension to match the sizes corresponding to different scale style features, which are concatenated layer by layer to obtain the watermarked style features, as follows:
\begin{eqnarray}\label{2}
f_{wm}^{sty}= \mathcal{F}_{sty}(I_{sty},w_{ori}).
\end{eqnarray}

Adaptive affine transformations regulate image style via the generator network's normalization statistics, commonly used in AST for feature alignment. This step crucially disrupts watermark distributions during actual AST. We hypothesize that if the watermark encoder learns the AST feature transfer process and uses it to reconstruct watermarked images, AST will transfer the watermark with style features to natural images similarly. Based on this, we use adaptive normalization to compute affine parameters from watermarked style features and align content feature to obtain stylized intermediate layer features with watermark traces:
\begin{eqnarray}\label{3}
f_{wm}=\sigma  (f_{wm}^{sty})(\frac{f_{con}-\mu(f_{con})}{\sigma (f_{con})} )+\mu (f_{wm}^{sty}),
\end{eqnarray}
where  $f_{wm}$ denotes stylized features, $\mu$ , $\sigma$ are the mean and standard deviation. This hypothesis is validated in experiments. As shown in Figure~\ref{fig3}, StyleMark's watermark residuals ($R_{wm}$) primarily capture varying degrees of brushstroke features rather than outlining the image's structural contours. Post-AST residual images ($R_{wm}^{cs}$) demonstrate effective transfer of embedded watermarks from style to semantic structure contours of natural images.


$f_{wm}$ is fed into the upsampling decoder $\mathcal{F}_{dec}(\cdot)$ to reconstruct the watermarked style image $I_{wm}$. Direct image reconstruction after feature alignment may lose content structure details. To mitigate this and prevent overfitting from deep network layers, we introduce residual connections with the original image during reconstruction and re-embed watermark information to prevent overwriting, as follows:
\begin{eqnarray}\label{4}
I_{wm}=\mathcal{F} _{dec}(f_{wm},I_{sty},w_{ori}).
\end{eqnarray}

\subsubsection{Invisibility Loss.}The watermark encoding must not affect the perceived quality of the style image or the normal use of authorized AST models. We use MSE loss to constrain distortion in the overall color and brightness of the watermarked style image, ensuring the invisibility of the watermark information:
\begin{eqnarray}\label{mse}
\mathcal{L} _{mse}=MSE(I_{wm},I_{sty}).
\end{eqnarray}

Furthermore, during image reconstruction, part of the watermark information may be distributed into content features, which cannot be preserved after AST. Therefore, To fully utilize the redundant style features, we propose a distribution squeeze loss $\mathcal{L}_{dsl}$. This optimizes intrinsic watermark distribution by compressing normalized feature differences of the content structure in the grayscale domain, forcing the encoder to embed most of the watermark into style features. We first extract grayscale image content features using the content feature encoder:  
\begin{eqnarray}\label{6}
f_{wm}^{gray}, f_{ori}^{gray}=\mathcal{F} _{con}(gray(I_{wm}),gray(I_{sty})),
\end{eqnarray}
where $gray(\cdot)$ denotes the operation of converting to grayscale. $\mathcal{L}_{dsl}$ is then computed using the MSE loss after normalizing the grayscale content features:
\begin{eqnarray}\label{gray}
\mathcal{L} _{dsl}=MSE(norm(f_{wm}^{gray}),norm(f_{ori}^{gray})),
\end{eqnarray}
where $norm(\cdot)$ represents the normalization operation. The overall invisibility loss $\mathcal{L}_{inv}$ consists of $\mathcal{L} _{mse}$ and $\mathcal{L} _{dsl}$:
\begin{eqnarray}\label{img}
\mathcal{L} _{inv}=\lambda _{mse}\mathcal{L} _{mse}+\lambda _{dsl}\mathcal{L} _{dsl},
\end{eqnarray}
where $\lambda _{mse}$ and $\lambda _{dsl}$ are hyper-parameters balancing terms.
\subsection{StyleMark Decoder Optimization}

\subsubsection{Watermark Decoding.}The watermark decoder must be able to extract watermarks at various stages of the watermarked image's dissemination for image attribution, requiring robustness against various levels of distortion. However, optimizing for robustness conflicts with invisibility, as increased robustness leads to greater degradation of the carrier image quality. Therefore, in the first training stage, we perform end-to-end training of the encoder and decoder without introducing the pixel-level distortion noise pool in the feedforward process, and pre-train the decoder with AST distortions. In the second-stage, we freeze the encoder parameters, introduce random noise to each mini-batch, and fine-tune the decoder. We Use ResNet50~\cite{he2016deep} as the backbone of the StyleMark watermark decoder, with a sigmoid layer for binary probability prediction. During the first-stage, the decoder decodes the watermarked style image $I_{wm}$ to extract the watermark information $w_{sty}$ as follows:
\begin{eqnarray}\label{9}
w_{sty}=\mathcal{D} (I_{wm}).
\end{eqnarray}

Unlike general local image editing processes or pixel-level distortions, AST does not retain the structural and semantic information of the style image. This irreplaceability motivates us to link the AST process before the decoder as a simulated noise pool for robustness training. The simplicity and effectiveness of AdaIN as the first AST method makes it the baseline for most subsequent ST studies~\cite{huang2017arbitrary}. It can be seamlessly integrated into our StyleMark training framework, ensuring computational efficiency under multi-model parallelism. More importantly, simple distortion simulation often leads to more generalized robustness enhancement. Therefore, we use the AdaIN model as the target model for white-box robustness training. The watermarked style image $I_{wm}^{sty}$ and content image $I_{con}$ are fed into the AST model to obtain the watermarked stylized image $I_{wm}^{cs} $:
\begin{eqnarray}\label{10}
I_{wm}^{cs} =\mathcal{AST} (I_{wm},I_{con}).
\end{eqnarray}

The decoder learns to decode the watermark information $w_{cs}$ from the stylized image:
\begin{eqnarray}\label{11}
w_{cs} =\mathcal{D} (I_{wm}^{cs}).
\end{eqnarray}


\subsubsection{Watermark Loss and Two-Stage Loss.}We use the BCE loss to calculate the difference between the recovered watermark and the original watermark. The watermark information loss in first-stage is as follows:
\begin{eqnarray}\label{12}
\mathcal{L} _{wm}^{stage1} =BCE(w_{sty},w_{ori})+BCE(w_{cs},w_{ori}).
\end{eqnarray}

The total loss in first-stage can be expressed as follows:
\begin{eqnarray}\label{13}
\mathcal{L} _{stage1} =\lambda_{wm}  \mathcal{L} _{wm}^{stage1} +\lambda_{inv} \mathcal{L} _{inv},
\end{eqnarray}
where $\lambda_{wm}$ and $\lambda_{inv}$ denote different loss weights.

In the second fine-tuning stage, we consider a wider range of common pixel-level distortions. Artworks may be subjected to image processing operations imposed by channels or platforms during dissemination. Additionally, AST-generated images may also undergo image processing or malicious post-processing attempts to erase watermark traces. Therefore, we construct a pixel-level distortion noise pool $\mathcal{NP}(\cdot)$ that includes Gaussian noise, JPEG compression, and Gaussian blur. For each mini-batch, we randomly select a noise to process both the watermarked style image $I_{wm}$ and the stylized image $I_{wm}^{cs}$. The decoder decode watermarks from images under different distortions, as follows:
\begin{eqnarray}\label{11}
w_{sty} =\mathcal{D} (I_{wm}), \hat{w}_{sty} =\mathcal{D} (rand(\mathcal{NP}(I_{wm}))),
\end{eqnarray}
where $rand(\cdot)$ denotes a random selection operation. The training loss for the second-stage is defined as follows:
\begin{align}\label{16}
\begin{split}
\mathcal{L} _{stage2} &=BCE(w_{sty},w_{ori})+BCE(w_{cs},w_{ori}) \\
&+BCE(\hat{w}_{sty},w_{ori})+BCE( \hat{w}_{cs},w_{ori}).
\end{split}
\end{align}

\section{Experiments}
\subsection{Implementation Details}

\begin{figure*}
	\centering
	\includegraphics[width=152mm]{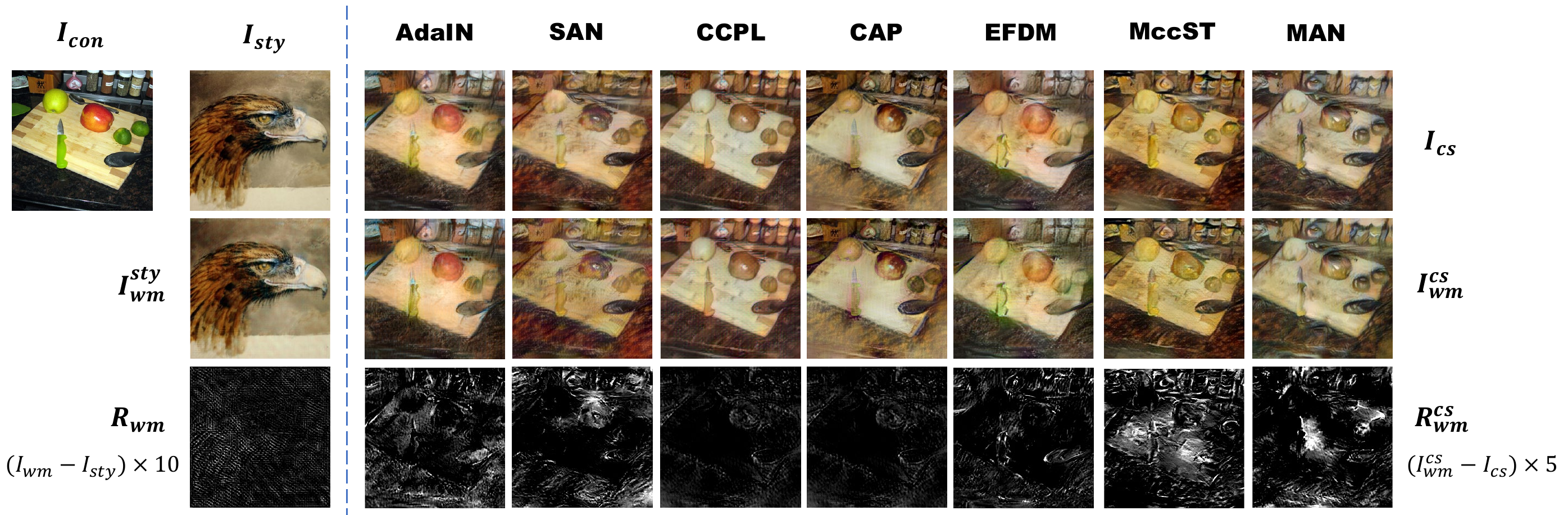}
        \caption{Subjective visual quality under various AST distortions. $R_{wm}$ represents the amplified residual image of $I_{wm}$ and $I_{sty}$: $(I_{wm}-I_{sty})\times 10$. $R_{wm}^{cs}$ represents the amplified residual image of $I_{wm}^{cs}$ and $I_{cs}$: $(I_{wm}^{cs}-I_{cs})\times 5$.}

 \label{fig3}
\end{figure*}
\subsubsection{Dataset:} The experiments conduct on the MSCOCO~\cite{lin2014microsoft} content image dataset and the WikiART~\cite{phillips2011wiki} style image dataset, as most open-source pre-trained AST models use these authoritative datasets. Specifically, 2000 pairs of style and content images are used for training, 500 pairs for validation, and 500 pairs for testing. Unless otherwise specified, all images are resized to $256\times256$ to match the dimensions of the pre-trained AST models. The default watermark information length is set to 30 bits, which can represent over a billion different copyright attributions, sufficient for practical requirements. 

\subsubsection{Implementation Details:} StyleMark is implemented by PyTorch and executed on NVIDIA RTX 3090. In Alg.~\ref{alg}, $Stage1\_iter$ and $Stage2\_iter$ are set to 4000 and 200. The batch size is 16. We empirically adjust the Adam optimizer with an initial learning rate of 0.00001. In Eq.~\ref{img} and Eq.~\ref{13}, $\lambda_{mse}$, $\lambda_{dsl}$, $\lambda_{inv}$, $\lambda_{wm}$  are empirically  set to 1, 0.2, 1, 0.002. We use seven SOTA AST methods for evaluation, including AdaIN~\cite{huang2017arbitrary}, SANet~\cite{park2019arbitrary}, CCPL~\cite{wu2022ccpl}, EFDM~\cite{zhang2022exact}, MccSTN~\cite{mccstn}, MANet~\cite{deng2020arbitrary}, CAP~\cite{cap}, all using their officially provided open-source pre-trained models to ensure resulting convincing artistic styles. Among them, the AdaIN model is used as the white-box model in training, while the others are black-box models used solely for evaluation. StyleMark model structure details are provided in the Appendix.
\begin{table*}
	\belowrulesep=0pt
       \aboverulesep=0pt
	\renewcommand\arraystretch{1.2}  
\resizebox{\hsize}{!}{
 
       \begin{tabular}{ccc|c|ccccccccccccccc|c}
       \Xhline{1px}
       \multirow{3}{*}{\textbf{Method}} & \multicolumn{2}{c|}{\textbf{Invisibility}}& \multicolumn{17}{c}{\textbf{Robustness}}\\ 
       \cmidrule(r){2-20}
       & \multirow{2}{*}{\textbf{PSNR}}& \multirow{2}{*}{\textbf{SSIM}}
      & \multirow{2}{*}{\textbf{Ori}}&\multicolumn{8}{c}{\textbf{Robustness to black-box AST}} & \multicolumn{7}{c|}{\textbf{Robustness to common distortion }}&\multirow{2}{*}{\textbf{Average}}\\
    \cmidrule(r){5-11}\cmidrule(r){12-19}
			
			& &  & &AdaIN  &SANet  &CCPL  &CAP  &EFDM  &MANet  &MccSTN & \makecell[c]{Brightness\\($f=0.5$)}  &\makecell[c]{Contrast\\($f=0.5$)} &\makecell[c]{Hue\\($f=0.1$)}  &\makecell[c]{GaussianNoise\\($\sigma=0.005$)} &\makecell[c]{GaussianBlur\\($0.003,30$)} &\makecell[c]{JPEG\\($QF=50$)} & \makecell[c]{Resize\\($p=50\%$)}  & \makecell[c]{Saturation\\($f=0.5$)}\\
   \cmidrule(r){1-11}\cmidrule(r){12-20}
			HiDDeN\shortcite{zhu2018hidden} &33.492 &0.926 &\textbf{1} &0.498 &0.476 &0.501 &0.492 &0.498 &0.500 &0.489 &0.971 &0.978 &0.974 &0.921 &0.989 &0.954 & 0.980 &\textbf{1} &0.764\\
		      MBRS*\shortcite{jia2021mbrs}  &34.621 &0.902 &\textbf{1} &0.502 &0.492 &0.486 &0.475 &0.501 &0.493 &0.500 &\textbf{1} &\textbf{1} &\textbf{1} &0.985 &0.957 &0.948 &0.978 &\textbf{1} &0.770 \\
              MBRS\shortcite{jia2021mbrs}  &35.271 &0.949 &\textbf{1} &0.504 &0.500 &0.493 &0.499 &0.509 &0.489 &0.487 &\underline{0.979} &\textbf{1} &\textbf{1} &\textbf{1} &0.976 &\textbf{1} &0.982 &\textbf{1} &\underline{0.776}\\
            FakeTagger\shortcite{wang2021faketagger} &\textbf{38.512} & \underline{0.973} &0.993 &0.506	&0.500	&0.472	&0.501	&0.473	&0.495 &0.497 &0.843  &0.956 &0.970 &0.957 &0.782 &0.800 &0.725 &0.970 &0.715\\
		PIMoG*\shortcite{fang2022pimog}  &\underline{37.821} &0.944 &\underline{0.998} &0.492	&0.501	&0.477	&0.498	&0.496	&0.493	&0.488  &0.957 &0.997 &0.986 &0.988 &0.982 &0.952 &0.965 &\underline{0.998} &0.767\\
              PIMoG\shortcite{fang2022pimog}  &33.033 &0.943 &0.989 &0.496	&0.498	&0.487	&0.506	&0.492	&0.483	&0.486  &0.965 &0.984 &0.952 &0.986 &0.802 &0.975 &0.892 &0.978  &0.717\\
               FIN\shortcite{fang2023flow}  &32.668 &\textbf{0.983} &1 &0.498	&0.486	&0.501	&0.492	&0.502	&0.496	&0.500  &0.939 &\underline{0.998} &\underline{0.995} &\textbf{1} &\underline{0.992} &\textbf{1} &\textbf{1} &\underline{0.998} &0.774\\
		   \textbf{StyleMark(Ours)}  &35.632 &0.952 &\textbf{1} &\textbf{0.986} &\textbf{0.884} &\textbf{0.821} &\textbf{0.836} &\textbf{0.904}&\textbf{0.887} &\textbf{0.852} &0.967 &\textbf{1} &0.977 &\underline{0.994} &\textbf{0.997}&\underline{0.983}&\underline{0.986} &\textbf{1} &\textbf{0.942}\\
			\hline

\Xhline{1px}
	\end{tabular}}
 \caption{Invisibility and robustness evaluation. The best results are in bold and sub-optimal results are underlined. }
	\label{t1}

\end{table*}


\subsubsection{Baselines:} We compare StyleMark with deep robust watermarking methods, including HiDDeN~\cite{zhu2018hidden}, MBRS~\cite{jia2021mbrs}, FakeTagger~\cite{wang2021faketagger}, PIMoG~\cite{fang2022pimog} and FIN~\cite{fang2023flow}. For a fair comparison, we use open-source code of the aforementioned methods, training on the same image size and watermark length as StyleMark, following the procedures in their papers. Training details are in the Appendix. MBRS and PIMoG pre-trained parameters are open-sourced for $128\times128$ images. For rigorous evaluation, we include these test results, marked by $\ast$ in Table~\ref{t1}. 


\subsubsection{Metrics:} We use PSNR, SSIM~\cite{Wang_Bovik_Sheikh_Simoncelli_2004} to evaluate the quality of the watermarked style image, use the accuracy rate $Acc$ to evaluate the watermark recovery accuracy: $Acc=1-\frac{1}{L}\sum_{l=1}^{L}\left | w_{ori}- w_{*} \right |$,
where $w_{*}$ denotes the watermark recovered from different stages, $l$ and $L$ represent the bits index and the length of the watermark respectively.

\subsection{Robustness Evaluation}

We evaluate the robustness of StyleMark against various distortions, including black-box AST and common distortions. As shown in Table~\ref{t1}, after processing watermarked images with different black-box AST models, StyleMark consistently recovered the watermark with high accuracy. In contrast, other methods failed to withstand AST, with recovery accuracy close to random guessing. Additionally, robust watermarking algorithms for strong distortions often overfit specific image transformations. To validate StyleMark's generalization, we evaluate it against common pixel-level distortions. Across eight different image processing operations, StyleMark maintain robustness comparable to specialized methods, achieving optimal or near-optimal accuracy, exceeding $97\%$. The average accuracy in robustness evaluation demonstrates that StyleMark achieves optimal robustness compared to other SOTA methods.


\subsection{Security Evaluation}
\subsubsection{Post-Processing of Stylized Images.}


AST-generated counterfeit artworks may circulate online and undergo necessary or malicious post-processing, which can disrupt the watermark information. We subjected AST-generated stylized images to common distortions and evaluated the security of StyleMark. As shown in Table~\ref{t2}, even when $I_{wm}^{cs}$ is overlaid with various common distortions, the accuracy does not significantly decrease, with a maximum drop of less than $8\%$, ensuring substantial watermark security.

\subsubsection{Adaptive Attacks.} 

\textit{A. Regular adversary}: Art style plagiarists may use their open-source watermarking algorithms to embed fake watermarks into original artworks and then extract these fake watermarks from AST-generated images, claiming them as their own original art style. We evaluated the security  of StyleMark against such watermark rewriting attacks using FakeTagger and MBRS algorithms. StyleMark watermarks were first embedded in the original images, followed by the fake watermark ($w_{ft}$ or $w_{mbrs}$) using FakeTagger or MBRS. As shown in Table~\ref{t3} this type of watermark rewriting has minimal impact on the recovery accuracy of StyleMark, whereas FakeTagger and MBRS cannot resist AST, failing to recover corresponding watermarks.


 \begin{table}
	\belowrulesep=0pt
       \aboverulesep=0pt
	\renewcommand\arraystretch{1.2}  
\resizebox{\hsize}{!}{
 
       \begin{tabular}{c|c|c|c|c|c|c|c}
       \Xhline{1px}

			Distortion  &AdaIN  &SANet  &CCPL  &CAP   &EFDM  &MANet  &MccSTN\\
            \hline
			ori &0.986 &0.884 &0.821 &0.836  &0.904 &0.887 &0.852 \\
			Brightness  &0.902	&0.805	&0.801	&0.802	&0.820 &0.811	&0.774 \\
                Contrast  &0.950	&0.867	&0.809	&0.831	&0.892	&0.856	&0.801\\
                Hue &0.933	&0.860&0.812	&0.816	&0.868	&0.847	&0.803\\
			GaussianNoise&0.952	&0.846	&0.823	&0.801 &0.872	&0.872&0.831\\
                GaussianBlur&0.953	&0.860	&0.811	&0.816	&0.884	&0.855	&0.832\\
			JPEG  &0.950	&0.863	&0.803	&0.842	&0.898	&0.853	&0.808\\
   Resize  &0.907	&0.834	&0.806	&0.801	&0.860	&0.832	&0.788\\
   Saturation  &0.956	&0.847	&0.815	&0.820	&0.883	&0.862	&0.816\\
             \hline
    Average  &\makecell[c]{0.943\\(($\downarrow$\textbf{0.043}) )} &\makecell[c]{0.852\\($\downarrow$\textbf{0.032 )}} &\makecell[c]{0.811\\($\downarrow$\textbf{0.010})}&\makecell[c]{0.818\\($\downarrow$\textbf{0.018} )}&\makecell[c]{0.876\\($\downarrow$\textbf{0.028} )} &\makecell[c]{0.853\\($\downarrow$\textbf{0.034} )} &\makecell[c]{0.812\\($\downarrow$\textbf{0.040} )}\\
   
\Xhline{1px}
	\end{tabular}}
 \caption{Security evaluation of post-processing. $\downarrow$ indicates performance degradation compared to the original scenario.}
	\label{t2}

\end{table}

\textit{B. Professional adversary}: Although the StyleMark model is not accessible, a knowledgeable adversary could potentially use the StyleMark training framework to train their own watermark models. We retrained MBRS and FakeTagger models using the StyleMark framework (MBRS$\ast$  and FakeTagger$\ast$ ). As shown in table~\ref{t3}, both retrained methods fail to recover watermarks after AST. This is because the models embed watermarks directly into the global feature space, and AST disrupts the watermark distribution within the style image's content, hindering effective training. StyleMark can recover watermarks with high accuracy even after multiple overlays, ensuring secure traceability.


\begin{table}
	
	\renewcommand\arraystretch{1.2}  
\resizebox{\hsize}{!}{
 
       \begin{tabular}{c|c|c|c|c|c|c|c}
       \Xhline{1px}

			Distortion  &AdaIN  &SANet  &CCPL  &CAP  &EFDM  &MANet  &MccSTN\\
            \hline
			FakeTagger ($w_{ft}$)&0.501 &0.498 &0.503 &0.487 &0.502&0.487 &0.492 \\
    \textbf{StyleMark ($w_{ori}$)} & 0.964	&0.871	&0.819	&0.831	&0.900	&0.882	&0.836
 \\  
  \hline
     \rowcolor{gray!20}  FakeTagger* ($w_{ft}$) & 0.497	&0.504	&0.486	&0.493	&0.481	&0.483	&0.496\\
 \rowcolor{gray!20}   \textbf{StyleMark ($w_{ori}$)} & 0.972	&0.879	&0.810	&0.827	&0.895	&0.880	&0.842
 \\    
  \hline
			MBRS ($w_{mbrs}$) &0.497	&0.492	&0.506	&0.491	&0.493 &0.487	&0.500 \\
			
			\textbf{StyleMark ($w_{ori}$)} & 0.960	&0.862	&0.809	&0.825	&0.889	&0.873	&0.829
 \\
  \hline
  \rowcolor{gray!20}  MBRS* ($w_{mbrs}$) & 0.495	&0.503	&0.487	&0.476	&0.506	&0.483	&0.489\\
\rowcolor{gray!20}   \textbf{StyleMark ($w_{ori}$)} & 0.971	&0.868	&0.807	&0.819	&0.879	&0.870	&0.833
 \\
			\hline

\Xhline{1px}
	\end{tabular}}
 \caption{Security evaluation against watermark overwriting attacks. Gray-shaded rows indicate that the model was trained using the StyleMark training framework.}
	\label{t3}

\end{table}

\subsection{Image Visual Quality}


In Figure~\ref{fig3}, we present the visual effects of StyleMark. It is evident that watermark embedding does not cause visual degradation to host artworks, with no perceptible artifacts in $I_{wm}$. StyleMark does not affect the normal AST generation, as there is no significant visual difference between $I_{wm}^{cs}$ and the original AST-generated image $I_{cs}$. From $R_{wm}$, it is evident that StyleMark embeds the watermark within irregular brushstroke features, independent of structural conceptual information. $R_{wm}^{cs}$ demonstrates that AST learns the watermark traces along with the brushstroke features into natural content images. This validates the motivation behind the StyleMark. Table~\ref{t1} presents the quantitative evaluation of invisibility. StyleMark achieves comparable invisibility to other methods, surpassing most. FakeTagger achieves impressive visual quality, but like all ohter methods, it fails to recover watermark after AST, similar to random guessing ($\sim 50\%$ $Acc$) and lacking  robust against common distortions. More visualizations are shown in the Appendix.

\subsection{Ablation Study}

\begin{figure}
	\centering
	\includegraphics[width=75mm]{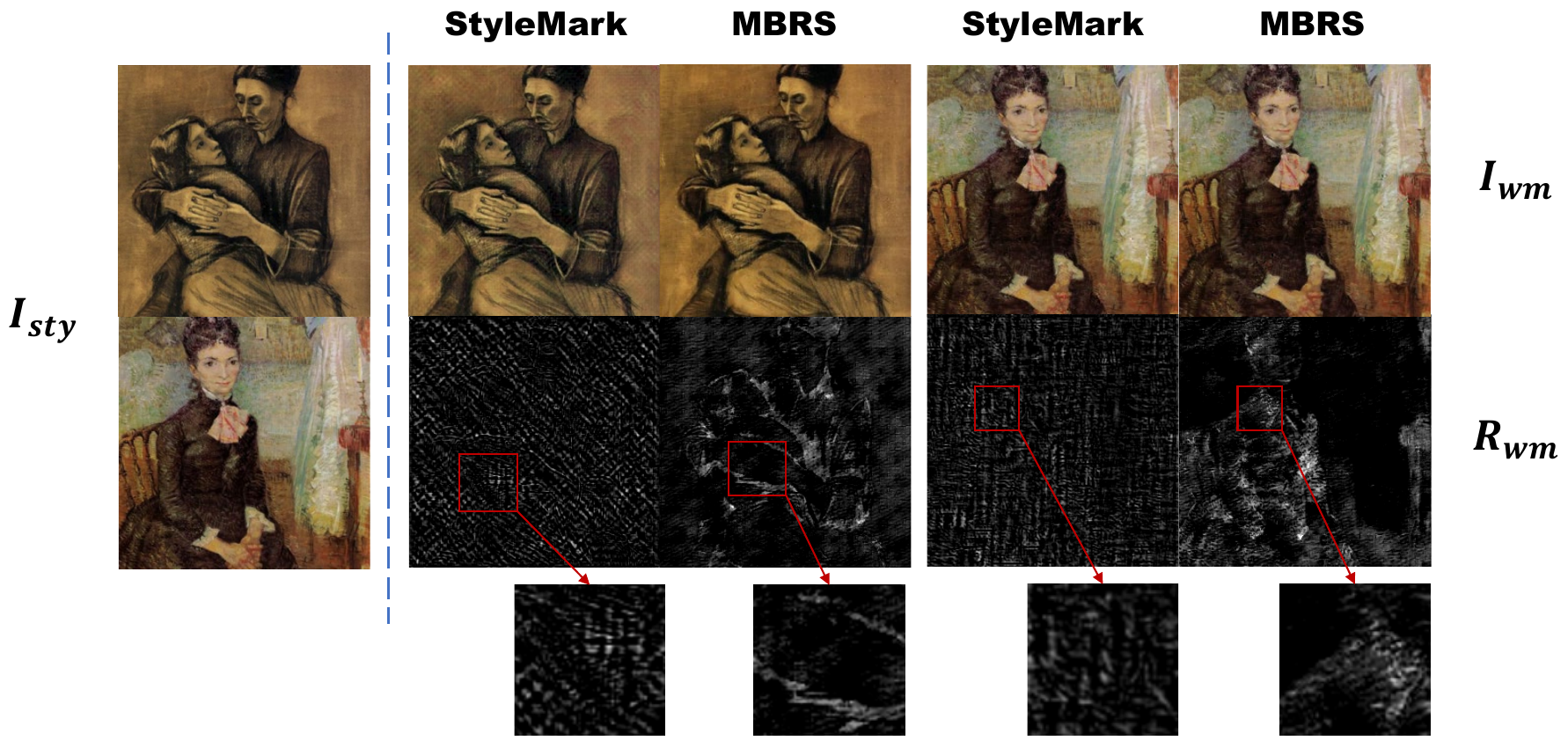}
        \caption{Visualization of images from different models.}

 \label{fig4}
\end{figure}

\begin{table}
	\belowrulesep=0pt
       \aboverulesep=0pt
	\renewcommand\arraystretch{1.2}  
\resizebox{\hsize}{!}{
 
       \begin{tabular}{c|c|ccccc}
       \Xhline{1px}

			& &  w/o.$\mathcal{L} _{dsl}$ &  w/o.residual  &w/o.network  &StyleMark \\
            \hline
			\multirow{2}{*}{\textbf{Invisibility}} &PSNR &30.236 ($\downarrow$ \textbf{5.396}) &23.040 ($\downarrow$ \textbf{12.592}) &33.674 ($\downarrow$ \textbf{1.958}) &35.632\\
			 & SSIM &0.853 ($\downarrow$ \textbf{0.099}) &0.650 ($\downarrow$ \textbf{0.302}) &0.923 ($\downarrow$ \textbf{0.029}) &0.952\\
          \hline
                \multirow{9}{*}{\textbf{Robustness}} & AdaIN &0.974 &0.922 &0.492 &0.986\\
          
			& SANet &0.853 &0.798 &0.502 &0.884 \\
			&CCPL &0.800 &0.795 &0.500 &0.821\\
   &CAP &0.801 &0.792  &0.475 &0.836\\
   &EFDM &0.882 &0.860 &0.501 &0.904\\
   &MANet &0.852 &0.879 &0.482 &0.887\\
   &MccSTN &0.827 &0.801 &0.486 &0.852\\
    \cmidrule(r){2-6}
   &Average &0.856 ($\downarrow$\textbf{0.025}) &0.835 ($\downarrow$ \textbf{0.046}) &0.491 ($\downarrow$ \textbf{0.390}) &0.881\\
			\hline

\Xhline{1px}
	\end{tabular}}
 \caption{Ablation experiments of various method variants.}
	\label{t3}

\end{table}

\subsubsection{Importance of Watermark Model and Its Related Structure.} To validate the importance of the proposed watermark model, we retained the StyleMark training framework but replaced its encoder and decoder with those from MBRS, followed by end-to-end training and decoder fine-tuning. As shown in Table~\ref{t3} (w/o. network), the MBRS encoder embeds watermarks in the global feature space. Even with AST distortions during training, the watermarks couldn't be accurately recovered, similar to random guessing. We visualize the differences caused by different models in watermarked images. As shown in the residual images in Figure~\ref{fig4}, the MBRS encoder embeds most watermark information into the image's structural contours, which is lost along with the content information after AST. In contrast, StyleMark embeds watermarks into the texture and unique brushstroke features, facilitating synchronous learning of watermark traces by the AST model. Additionally, we introduce residual connections of the original style image into the image reconstruction process and re-embed multi-scale watermark information, instead of directly reconstructing the watermarked image from the aligned intermediate features. As shown in Table~\ref{t3} (w/o. residual), This approach enhances detail recovery in watermarked images and improves robustness.



 \begin{table}
	\belowrulesep=0pt
       \aboverulesep=0pt
	\renewcommand\arraystretch{1.2}  
\resizebox{\hsize}{!}{
 
       \begin{tabular}{c|c|c|c|c|c|c|c|c}
       \Xhline{1px}

			Distortion  &Ori &AdaIN  &SANet  &CCPL  &CAP  &EFDM  &MANet  &MccSTN\\
            \hline
			Ori &1 &0.961 &0.836 &0.821 &0.792 &0.884 &0.857 &0.825 \\
			Brightness &0.908 &0.833	&0.795	&0.732	&0.754	&0.812 &0.779	&0.764 \\
                Contrast &0.982 &0.933	&0.809	&0.768	&0.796	&0.872	&0.842	&0.768\\
                Hue &0.960 &0.905	&0.803 &0.761	&0.766	&0.838	&0.815	&0.736\\
			GaussianNoise&0.856 &0.698	&0.636	&0.667	&0.679 &0.621	&0.609 &0.623\\
                GaussianBlur&0.840 &0.752	&0.675	&0.689	&0.648	&0.701	&0.700	&0.671\\
			JPEG  &0.820 &0.830	&0.738	&0.726	&0.712	&0.773	&0.760	&0.722\\
   Resize &0.796  &0.723	&0.687	&0.629	&0.632	&0.684	&0.652	&0.638\\
   Saturation &0.973  &0.956	&0.825	&0.758	&0.786	&0.857	&0.823	&0.757\\
   \hline
   \multirow{2}{*}{\textbf{Average}}  &0.903  &0.843	&0.756	&0.727	&0.729	&0.784	&0.760	&0.722\\
   &($\downarrow$ \textbf{0.086})  &($\downarrow$ \textbf{0.100})	&($\downarrow$ \textbf{0.096})	&($\downarrow$ \textbf{0.084})	&($\downarrow$ \textbf{0.089})	&($\downarrow$ \textbf{0.092})	&($\downarrow$ \textbf{0.093})	&($\downarrow$ \textbf{0.090})\\

\Xhline{1px}
	\end{tabular}}
 \caption{Ablation experiments without fine-tuning stage. The reduction is calculated by comparing the data in Table~\ref{t2}.}
	\label{t5}

\end{table}
\subsubsection{Effectiveness of Distribution Squeeze Loss $\mathcal{L}_{dsl}$.} As shown in Table~\ref{t3}, removing $\mathcal{L}_{dsl}$ results in a decline in both invisibility and robustness. This indicates that watermarks in content features of style images are lost during the AST process. However, $\mathcal{L}_{dsl}$ forces the encoder to squeeze the watermarks into the style features by compressing content distortion, preserving watermark traces and aiding in the recovery of structural details in the watermarked image.


\begin{table}
	
	\renewcommand\arraystretch{1.2}  
\resizebox{\hsize}{!}{
 
       \begin{tabular}{ccccccccccc}
       \Xhline{1px}
       
			\multirow{2}{*}{\textbf{Length}} & \multicolumn{2}{c}{\textbf{Invisibility}}& \multicolumn{7}{c}{\textbf{Robustness}}\\ 
   \cmidrule(r){2-3}\cmidrule(r){4-10}
			 &PSNR &SSIM  &AdaIN  &SANet  &CCPL  &CAP  &EFDM  &MANet  &MccSTN\\
        \cmidrule(r){1-1}\cmidrule(r){2-3}\cmidrule(r){4-10}
			10 bits &37.552 &0.961 &0.994 &0.902 &0.865 &0.879 &0.923 &0.906 &0.879 \\
   
   15 bits &36.833 &0.958& 0.989	&0.906&0.859&0.852&0.919&0.901&0.873

 \\  
 
     30 bits &35.632 &0.952 & 0.986
	&0.884	&0.821	&0.836	&0.904	&0.887	&0.852\\
    
 40 bits &33.739 &0.939 &0.979 &0.893 &0.810 &0.816 &0.862 &0.859 &0.826

 \\    
  \hline

\Xhline{1px}
	\end{tabular}}
 \caption{Impact of watermark length on performance.}
	\label{t6}

\end{table}

\subsubsection{Influence of Fine-Tuning Stage.} To mitigate the conflict between invisibility and robustness during end-to-end training, we introduce a common pixel-level distortion noises pool and fine-tune the decoder for 200 epochs after initial pre-training. As shown in Table~\ref{t5}, thanks to solid pre-training, this simple fine-tuning stage significantly enhanced the robustness of the watermarke against post-processing, with accuracy improvements of up to $10\%$.

\subsubsection{Influence of Watermark Length.} We evaluated the impact of four different watermark lengths on performance, as shown in Table~\ref{t6}. Longer watermark sequences degrade image quality more, but even with 40-bit watermarks, StyleMark maintains over $81\%$ accuracy post-AST. Platforms can choose watermark lengths as needed. Since 30 bits can represent over a billion users, we set the default to 30 bits to ensure quality within a reasonable distortion range.

\section{Conclusion}


The new era of art sharing challenges the copyright protection of artworks. We propose StyleMark, the first copyright protection method for artworks targeting black-box AST. StyleMark embeds watermarks within style images to track their dissemination and verifies the style source after unauthorized AST. We propose a novel style watermark model for multi-scale watermark encoding in the shared style feature space, achieving watermark-accompanied style rendering. Additionally, we regularize global image loss with distribution squeeze loss to optimize internal watermark distribution. Finally, additional random noise fine-tuning enhances StyleMark's generalization. Extensive experiments demonstrates its robustness and security against various attacks. In future work, we will further optimize the balance between invisibility and robustness.  We except our research provides new insights for copyright protection in artworks.

\bibliography{aaai25}

\end{document}